\documentclass[letterpaper]{article} 
\usepackage{aaai24}  
\usepackage{times}  
\usepackage{helvet}  
\usepackage{courier}  
\usepackage[hyphens]{url}  
\usepackage{graphicx} 
\usepackage{natbib}  
\usepackage{caption} 
\usepackage{algorithm}
\usepackage{algorithmic}

\usepackage{multirow}
\usepackage{algorithm}
\usepackage{algorithmic}
\usepackage{multirow}
\usepackage{tabularx}
\usepackage{booktabs}
\usepackage{amsfonts,bm}
\usepackage{threeparttable}
\usepackage{xspace}
\usepackage{amsmath}
\usepackage{upgreek}
\usepackage{amssymb}
\usepackage{pifont}%
\usepackage{multirow}
\usepackage{makecell}
\usepackage{color}
\usepackage{graphicx}
\usepackage{colortbl}
\usepackage{soul}
\usepackage{subcaption}
\usepackage{wasysym}
\usepackage[first=0,last=9]{lcg}
\definecolor{Gray}{gray}{0.86}

\newcolumntype{x}[1]{>{\centering\arraybackslash}p{#1pt}}
\newcolumntype{y}[1]{>{\raggedright\arraybackslash}p{#1pt}}
\newcolumntype{z}[1]{>{\raggedleft\arraybackslash}p{#1pt}}
\usepackage{enumerate}
\usepackage[colorlinks]{hyperref}
\nocopyright

\usepackage{newfloat}
\usepackage{listings}
\DeclareCaptionStyle{ruled}{labelfont=normalfont,labelsep=colon,strut=off} 
\floatstyle{ruled}
\newfloat{listing}{tb}{lst}{}
\floatname{listing}{Listing}
\title{Spatial Transform Decoupling for Oriented Object Detection}
\author{
    Hongtian Yu\footnote{Equal contribution.},
    Yunjie Tian\footnotemark[1],
    Qixiang Ye,
    Yunfan Liu\footnote{Corresponding author.}
}
\affiliations{
    University of Chinese Academy of Sciences\\
    \{yuhongtian17, tianyunjie19\}@mails.ucas.ac.cn, \{qxye, liuyunfan\}@ucas.ac.cn
}

\begin{document}

\maketitle

\begin{abstract}
Vision Transformers (ViTs) have achieved remarkable success in computer vision tasks.
However, their potential in rotation-sensitive scenarios has not been fully explored, and this limitation may be inherently attributed to the lack of spatial invariance in the data-forwarding process.
In this study, we present a novel approach, termed Spatial Transform Decoupling (STD), providing a simple-yet-effective solution for oriented object detection with ViTs.
Built upon stacked ViT blocks, STD utilizes separate network branches to predict the position, size, and angle of bounding boxes, effectively harnessing the spatial transform potential of ViTs in a divide-and-conquer fashion.
Moreover, by aggregating cascaded activation masks (CAMs) computed upon the regressed parameters, STD gradually enhances features within regions of interest (RoIs), which complements the self-attention mechanism.
Without bells and whistles, STD achieves state-of-the-art performance on the benchmark datasets including DOTA-v1.0 (82.24\% mAP) and HRSC2016 (98.55\% mAP), which demonstrates the effectiveness of the proposed method.
Source code is available at \href{https://github.com/yuhongtian17/Spatial-Transform-Decoupling}{https://github.com/yuhongtian17/Spatial-Transform-Decoupling}.
\end{abstract}

\section{Introduction}

Recent years have witnessed substantial progress and notable breakthroughs in computer vision, which can be primarily attributed to the advent of Vision Transformer (ViT) models.
Benefiting from the powerful self-attention mechanism, ViTs consistently achieve new state-of-the-art performance across vision tasks including classification~\cite{vit, swin, hivit, fang2023eva}, object detection~\cite{vitdet, mimdet, itpn}, and semantic segmentation~\cite{xie2021segformer, hrformer}. 
Despite the progress made, the capability of ViTs in spatial transform invariance has not been fully explored and understood.
In many scenarios, ViTs are treated as a universal approximator, expected to automatically handle various vision data irrespective of their orientations and appearances.

\begin{figure}[!t]
    \centering
    \includegraphics[width=0.985\linewidth]{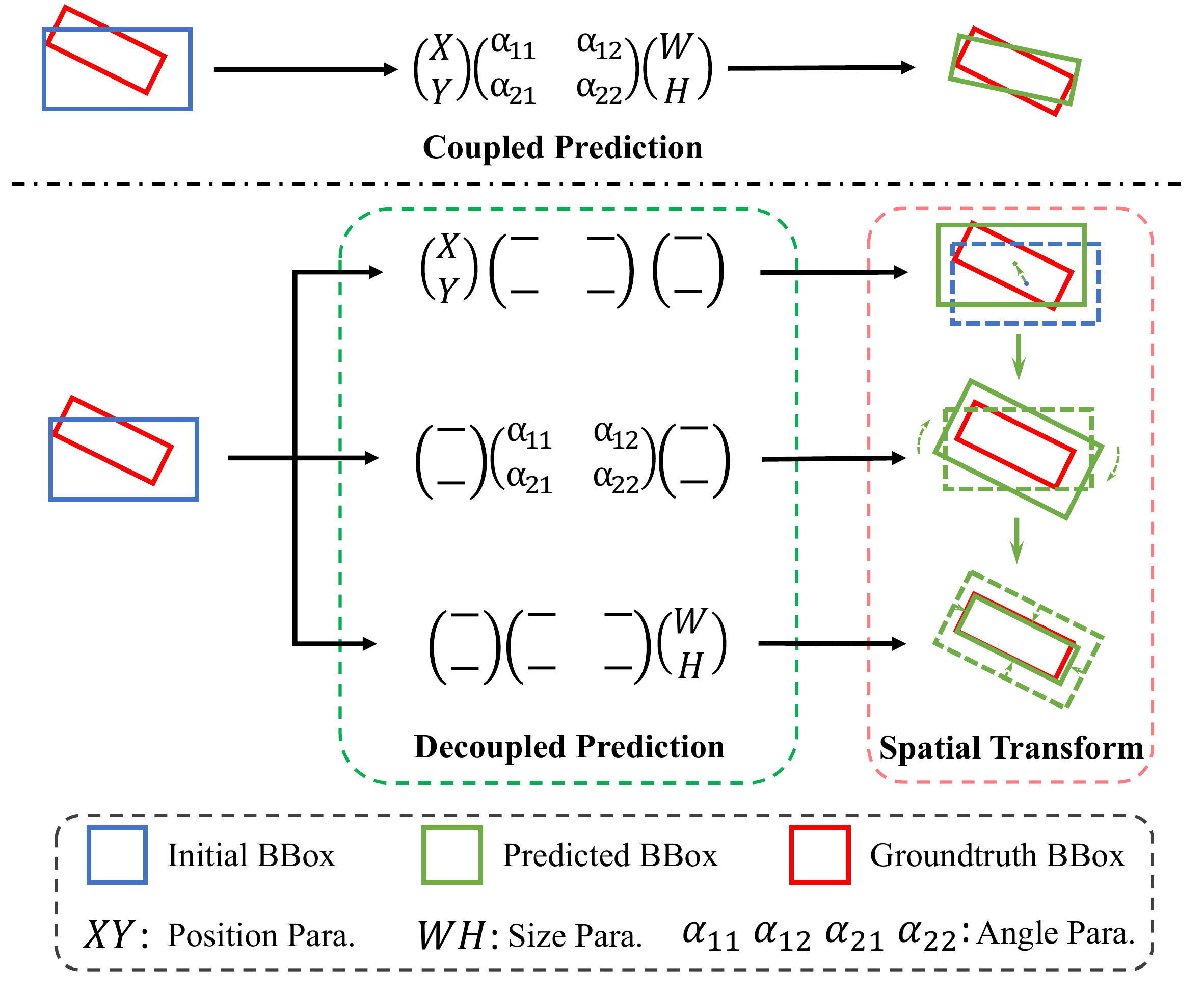}
    \caption{Conventional approaches (upper) estimate the position, size, and angle using a single RoI feature. In contrast, STD (lower) predicts and refines the parameters of bounding boxes in a divide-and-conquer (decoupled) manner.}
    \label{fig:mot}
\end{figure}

In this study, we aim to tap into the potential of ViTs in tackling the challenging spatial transform issue of vision tasks, $e.g.$, detecting objects in remote sensing scenarios, where images are captured from a bird's-eye view and target objects may appear in arbitrary orientations. 
To determine an oriented bounding box, initial research efforts~\cite{fasterrcnn, retinanet} suggested a direct regression approach for spatial transform parameters, including the spatial coordinates ($x$ and $y$), object size ($w$ and $h$), and the angle ($\alpha$).
However, such a straightforward regression strategy often results in discontinuous boundaries due to the inconsistency in angle representation and periodicity, as well as the suboptimal design of loss functions~\cite{csl, gwd, kld, kfiou}.

Rather than solely concentrating on developing more sophisticated angle representations or refining training objectives, it is essential to tackle the foundational issue of effectively extracting rotation-related features. 
In particular, we enhance the conventional structure of the bounding box prediction head by allocating distinct feature maps to predict parameters associated with diverse semantic interpretations, such as the object's location, shape, and orientation. 
This approach guides the feature extraction process in a controlled and effective manner.
Furthermore, by estimating the parameters associated with a particular spatial transform at each stage, this step-wise strategy facilitates the progressive refinement of estimation results, which in turn can contribute to improving the overall accuracy of the model.

Building upon the insights and discussions presented earlier, we propose Spatial Transform Decoupling (STD), a straightforward yet effective solution for oriented object detection, which decouples the estimation of transformation parameters related to object positions, sizes, and angles, Fig.~\ref{fig:mot}. 
Concretely, a multi-branch network design is utilized, where each individual branch is designated to predict parameters that correspond to distinct spatial transforms.
From another perspective, STD supplements the self-attention mechanism by allocating distinct responsibilities to self-attention modules at different stages of parameter prediction, which effectively utilizes the spatial transform capabilities of ViTs in a divide-and-conquer fashion.
Furthermore, STD integrates cascaded activation masks (CAMs) to enhance the features extracted by stacked Transformer blocks, effectively suppressing background information while highlighting foreground objects.
By refining features within regions of interest (RoIs) using CAMs, the feature representation for oriented objects is both decoupled and progressively enhanced.
As a simple-yet-effective design, STD can be integrated with various ViT-based detectors and achieve significant performance improvements over the state-of-the-art methods.
For instance, STD achieves 82.24\% mAP on DOTA-v1.0 and 98.55\% mAP on HRSC2016, surpassing the accuracy of all existing detectors.

The contributions of this work are summarized as:
\begin{itemize}
    \item The Spatial Transform Decoupling (STD) approach is introduced to address the challenge of oriented object detection by estimating parameters for spatial transforms through separate network branches. STD demonstrates remarkable generalizability and can seamlessly integrate with a variety of ViT detectors.
    
    \item Cascade activation masks (CAMs) are integrated into the self-attention module at each layer of ViT to progressively enhance the features. CAMs offer spatially dense guidance, directing the attention maps to focus more on foreground objects rather than the background.
    
    \item Experimental results demonstrate that STD surpasses state-of-the-art methods by a significant margin across a variety of oriented object detection benchmarks. 
\end{itemize}

\section{Related Work}

\subsection{Oriented Object Detection} 
Existing methods have investigated oriented object detection from the perspectives of feature robustness, region proposal refinement, and target regression enhancement.

\textbf{Feature Invariance/Equivalence.} 
Invariance or equivalence is an essential problem when designing/learning visual feature representations. During the era of hand-crafted features, SIFT~\cite{sift} utilizes dominant orientation-based feature alignment to achieve invariance to rotation and robustness to moderate perspective transforms. 
With the rise of CNNs, STN~\cite{stn} achieves rotation invariance by manipulating the feature maps according to the transformation matrix estimated using a sub-CNN. 
Group equivariant CNN~\cite{gcnn} proposes a natural generalization of CNNs, enabling them to group objects from the same categories regardless of orientations.
ORN~\cite{orn} introduces Active Rotating Filters (ARFs), which dynamically rotate during the convolution process and thereby produce feature maps with location and orientation explicitly encoded.
ReDet~\cite{redet} achieves rotation-equivariant convolution (e2cnn~\cite{e2cnn}) by incorporating a rotation-invariant backbone, which normalizes the spatial and orientational information of features.

\begin{figure*}[t]
    \centering
    \includegraphics[width=0.98\linewidth]{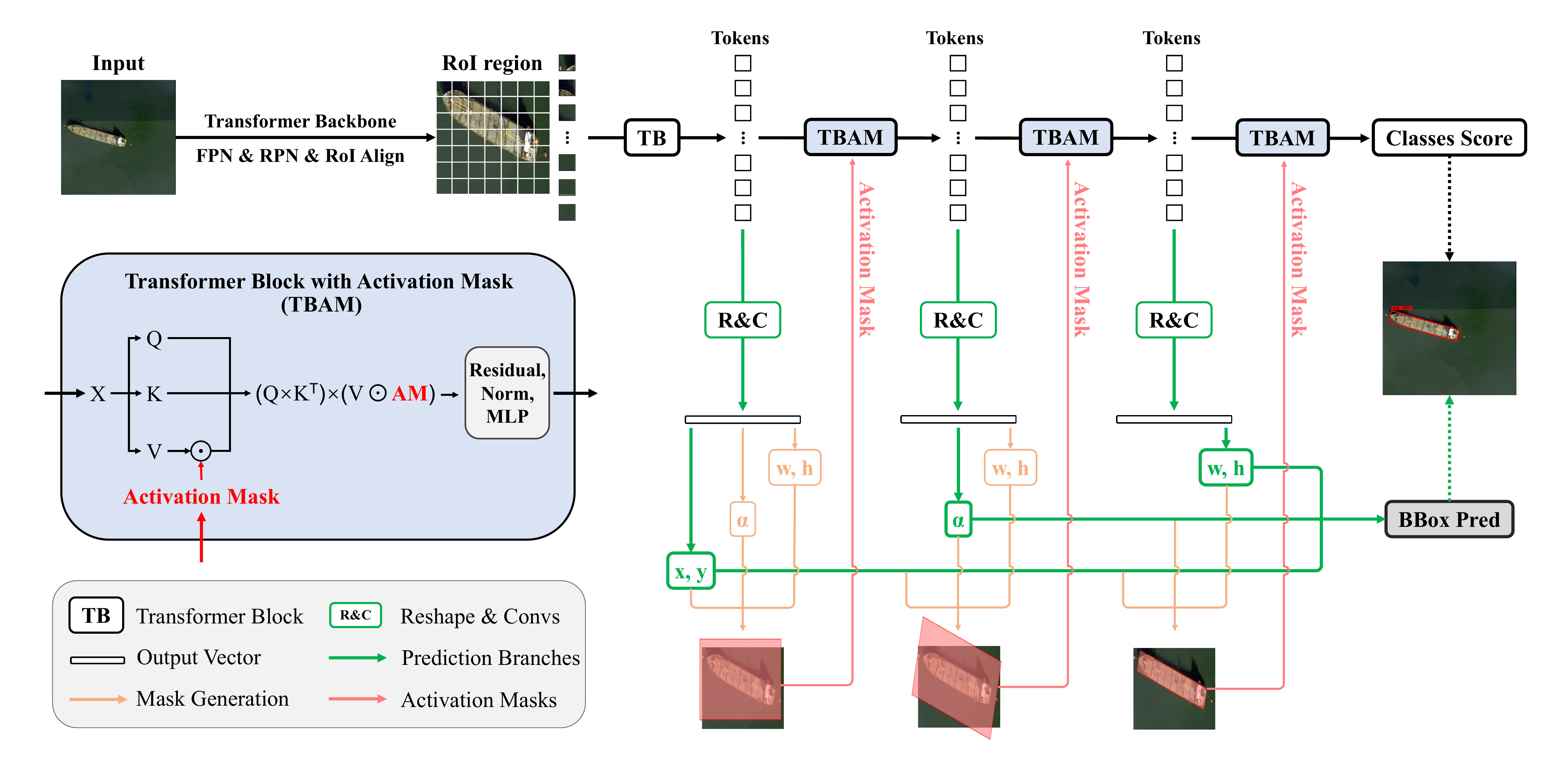}
    \caption{The framework of the proposed Spatial Transform Decoupling (STD) method. The detailed structure of Transformer blocks integrated with activation masks (TBAM) is shown on the left.}
    \label{fig:framework}
\end{figure*}

\textbf{Region Proposal Refinement.} 
RoI Transformer~\cite{roiTransformer} enhances two-stage detectors by iteratively repeating the RPN-RoI head structure~\cite{fasterrcnn, maskrcnn}.
Oriented RCNN~\cite{orientedrcnn} streamlines the process of oriented proposal generation and directly predicts oriented proposals based on the features extracted by the backbone and FPN (Feature Pyramid Network)~\cite{fpn} module.
Drawing inspiration from a similar concept, R$^{3}$Det~\cite{r3det} introduces a feature refinement stage to the orientation regression head.

\textbf{Target Regression Enhancement.} 
Gliding Vertex~\cite{glidingvertex} converts the task of rotated box prediction into regressing the offset for horizontal boxes along the four edges. 
CSL~\cite{csl} addresses the potential abrupt change in loss computation by proposing a label-based solution for angle prediction. 
CFA~\cite{cfa} and Oriented RepPoints~\cite{orientedreppoints} make improvements to the nine-point prediction methods~\cite{reppoints}. 
GWD~\cite{gwd}, KLD~\cite{kld}, and KFIoU~\cite{kfiou} use two-dimensional Gaussian distributions to solve the angle prediction problem. 
Despite the progress of various approaches proposed, few of them explore the impact of decoupling spatial transform, $e.g.$, position $(x, y)$, size $(w, h)$, and angle ($\alpha$), on the hierarchical feature representation.

\subsection{Vision Transformer} 
Drawing inspiration from the NLP field~\cite{attention, bert}, ViTs divide the image into multiple patch tokens for feature extraction and processing ~\cite{vit, swin, hivit, itpn}. 
It has attracted significant attention in recent years owing to its remarkable success in computer vision tasks.
DETR~\cite{detr} is a representative work that extends ViTs towards object detection, establishing the fundamental paradigm for applying ViT to this task. 
MAE~\cite{mae} proposes a novel pre-training mode that deviates from the classic fully supervised pre-training era of CNNs~\cite{cnnpretrain}. 
Building upon MAE, ViTDet~\cite{vitdet} and MIMDet~\cite{mimdet}, \textit{etc}, have made significant advancements in the development of ViT for object detection.

While Vision Transformers have demonstrated promising results in various visual tasks, they still encounter challenges in leveraging their advantages in handling object spatial transform, $e.g.$, oriented object detection.
Recently, RVSA~\cite{vsa, rvsa} made an initial attempt to improve the structure of ViT for oriented object detection tasks, which was achieved by updating Window Attention~\cite{swin, vitdet, mimdet} to Rotated Varied Size Attention.
Nevertheless, these methods solely rely on the self-attention mechanism to handle various spatial transformations, without explicitly introducing dense guiding information.

\section{The Proposed Method}

This section starts with an elucidation of the motivation behind Spatial Transform Decoupling (STD). 
Subsequently, a detailed explanation of the overall structure of STD is provided, offering an in-depth understanding of its architectural design and how it functions.
Next, we delve into a detailed decoupling structure and introduce the cascaded activation masks (CAMs) for progressive feature refinement.
Special emphasis is placed on their significant contribution to the overall performance enhancement of STD.

\subsection{Overview}

The proposed STD can be readily seen as an extension of existing oriented object detectors, and an overview of the architecture is depicted in Figure~\ref{fig:framework}. 
The primary innovation of STD resides within the detection head module, while for other components, such as the backbone, Region Proposal Network (RPN), and loss functions, we maintain consistency with mainstream detection frameworks~\cite{fasterrcnn, orientedrcnn}.
As a result, STD demonstrates significant generalizability, enabling its compatibility with a variety of detectors.
Specifically, for the purpose of a clear explanation, we adopt STD within the Faster RCNN framework~\cite{fasterrcnn} as the default configuration.
Throughout the experiments, we will also showcase the performance of STD in combination with other detectors, such as Oriented RCNN~\cite{orientedrcnn}.

ViTs have demonstrated impressive performance across a broad spectrum of visual tasks. However, their utilization in the context of oriented object detection remains relatively unexplored.
Nevertheless, existing pre-trained Transformer models are capable of extracting meaningful features, which contributes to establishing a strong foundation for achieving impressive performance in oriented object detection tasks.
Therefore, we adopt a design inspired by the imTED~\cite{imted} detector and substitute the backbone as well as head modules of the two-stage detector with \textbf{Vision Transformer blocks} pre-trained using the MAE method.

\begin{table}[!t]
\centering
\setlength{\tabcolsep}{0.09cm}
\begin{tabular}{@{}l|c|c|c@{}}
    & \multirow{2}*{2FCBBoxHead} & MAEBBoxHead & MAEBBoxHead \\
    & & (Not Pre-trained) & (Pre-trained) \\
\midrule
mAP & 69.67 & 69.16 & 71.07 \\
\end{tabular}
\caption{Performance comparison of Faster RCNN with the same backbone (ViT-small) but different heads. The training is carried out on the DOTA-v1.0 dataset~\cite{dota} for 12 epochs.}
\label{tab:baseline}
\end{table}

Specifically, we employ the ViT-small model as the backbone instead of ResNet-50, and use a 4-layer Transformer block to replace the conventional detection head in Faster RCNN built with fully connected (FC) layers.
Please note that the ViT-small backbone is obtained from the MAE pre-trained encoder, and the 4-layer Transformer block is derived from the pre-trained decoder, which forms the MAEBBoxHead module.
Once the regions of interest (RoIs) are obtained, the feature maps are uniformly divided into 7$\times$7 tokens, which are subsequently fed into the parameter regression head, as depicted in Figure~\ref{fig:framework}. 
Experiments are conducted to validate the effectiveness of this framework in addressing the oriented object detection problem, and the results are presented in Table~\ref{tab:baseline}.
In subsequent experiments, the pre-trained MAEBBoxHead is used as the baseline method by default.

Afterward, the proposed Spatial Transform Decoupling (STD) module is built upon the aforementioned backbone network.
To enhance the performance of decoupling, we employ a hierarchical structure to predict the bounding box parameters in a layer-wise manner, and further enhance it by leveraging the guidance provided by the cascaded activation masks (CAMs).
Detailed explanations of these contributions will be provided in the following two subsections.

\subsection{Decoupled Parameter Prediction}

As highlighted in the Introduction Section, different parameters of an oriented bounding box are expected to possess distinct properties (e.g., rotation-variance or rotation-invariance), and therefore, they should be computed based on different feature maps.
However, most conventional methods~\cite{fasterrcnn, retinanet} depend on a single feature map to predict all bounding box parameters, potentially resulting in the issue of coupled features.
To solve this problem, we introduce a multi-branch network to achieve hierarchical and disentangled parameter prediction.

As shown in Figure 2, we compute different components of the oriented bounding box based on the feature map at various stages of the Transformer decoder in a cascaded manner. 
Specifically, $\{x, y\}$, $\alpha$, $\{w, h\}$ and the class score are obtained based on the feature maps of the $1st$, $2nd$, $3rd$, and $4th$ layer of the Transformer block, respectively (the rationale behind this design will be detailed in ablation study).
After obtaining the discrete output from each Transformer block, we first reshape them into 7$\times$7 feature maps and then apply convolutional layers to further enhance the features. 
Next, after globally averaging the resultant feature maps, FC layers are adopted to make the final predictions, which are then used to produce the bounding box and CAMs (details explained in the next subsection).
Please note that the proposed mechanism is highly generalizable, as one can easily adjust the number of estimated parameters by simply adding or removing predicting branches.

\begin{figure}[!t]
    \centering
    \includegraphics[width=1.0\linewidth]{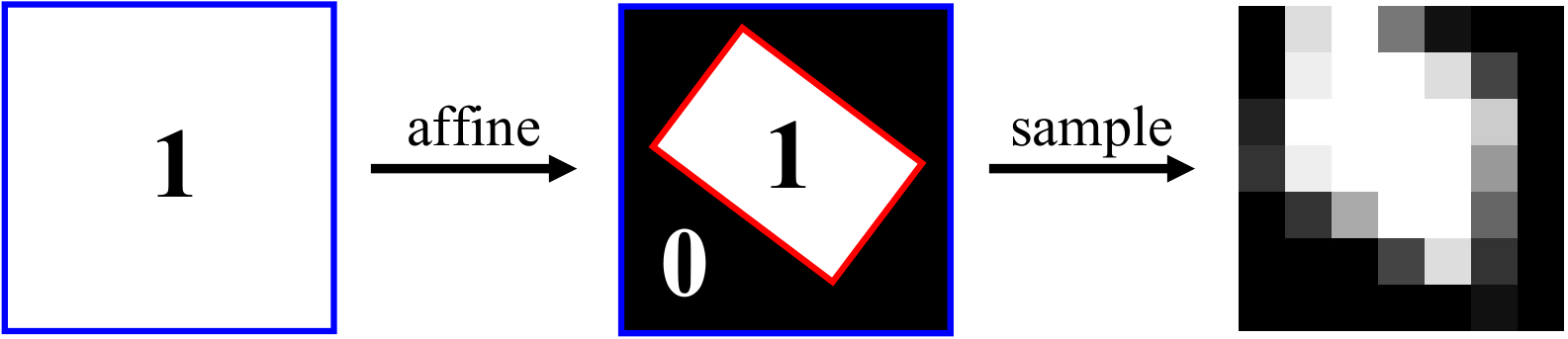}
    \caption{The translation between the predicted bounding box and the activation mask after affine transformation. The blue box represents the proposal region and the red box represents the activation mask.}
    \label{fig:actv_mask}
\end{figure}

\subsection{Cascaded Activation Masks}

To further regulate the decoupling process and improve the accuracy of prediction results, we intend to provide dense guidance for bounding box prediction at each stage.
To achieve this goal, cascaded activation masks (CAMs) are introduced to enhance the features generated by the multiple branches.

An ideal activation mask with binary values should have the regions corresponding to foreground objects assigned with a value of 1, and all background locations set to 0.
To align the activated regions with the foreground area as much as possible, we propose to generate activation masks by incorporating information from both the proposal and the predicted bounding box.

To be specific, the center point, size, and orientation of the estimated bounding box, $i.e.$, $(x_b, y_b, w_b, h_b, \alpha_b)$, could be expressed as 
\begin{equation}
    \begin{aligned}
        x_b      &= x_p + w_p \cdot dx,\\
        y_b      &= y_p + h_p \cdot dy,\\
        w_b      &= w_p \cdot e^{dw}, \\
        h_b      &= h_p \cdot e^{dh}, \\
        \alpha_b &= d\alpha
    \end{aligned}
\end{equation}
where $(x_p, y_p)$ and $(w_p, h_p)$ respectively denote the center coordinates and shape of the proposal, and $(dx, dy, dw, dh, d\alpha)$ are the predicted values related to the oriented bounding box obtained from STD.
Then, with the proposal placed in a rectangular coordinate system $(x, y)$ and its four vertices located at $(-1, -1)$, $(1, -1)$, $(1, 1)$, and $(-1, 1)$, the affine transformation against the bounding box $(x^{'}, y^{'})$ could be formulated as (please refer to ~\nameref{Appendix1}):
\begin{equation}
\label{eq:affine_transform}
    \resizebox{1.0\hsize}{!}{$%
    \left(\begin{array}{c}
                x^{'}  \\
                y^{'} 
            \end{array}\right)\\
    =\left(\begin{array}{cc}
                \cos d\alpha \cdot e^{dw} & -\sin d\alpha \cdot e^{dh} \cdot \frac{h_p}{w_p} \\
                \sin d\alpha \cdot e^{dw} \cdot \frac{w_p}{h_p} & \cos d\alpha \cdot e^{dh}
            \end{array}\right)
      \left(\begin{array}{c}
                x  \\
                y 
            \end{array}\right) \\
    +\left(\begin{array}{c}
                2\cdot dx \\
                2\cdot dy
            \end{array}\right) \\
    $}%
\end{equation}

As illustrated in Figure~\ref{fig:actv_mask}, the activation mask could be produced by applying the affine transformation in Eq.(\ref{eq:affine_transform}) to a matrix $\boldsymbol{AM}$ with all elements set to 1.
After integrating $\boldsymbol{AM}$ into the self-attention module in STD, the mapping function implemented by \textbf{Transformer Block with Activation Mask} (TBAM, see Figure~\ref{fig:framework}) could be written as 
\begin{equation}
    \text{TBAM}(\boldsymbol{Q}, \boldsymbol{K}, \boldsymbol{V}, \boldsymbol{AM}) = \text{softmax}(\frac{\boldsymbol{Q}\boldsymbol{K}^T}{\sqrt{d}})\boldsymbol{V'}
\end{equation}
where $\boldsymbol{V'}$ is obtained by performing an element-wise multiplication between $\boldsymbol{V}$ and $\boldsymbol{AM}$ ($\boldsymbol{V'}=\boldsymbol{V}\odot\boldsymbol{AM}$).
By multiplying with $\boldsymbol{V}$, $\boldsymbol{AM}$ could direct the model's attention by highlighting the foreground while suppressing the background.
In the forward propagation process, the utilization of activation maps guides the decoupled predicted values in earlier stages to direct the self-attention mechanism of the subsequent Transformer blocks; while during the backward propagation process, the discrepancies in the decoupled predicted values from later stages are propagated through the activation maps, affecting the feature extraction process of the previously decoupled predicted values.
This cascaded architecture enhances the interconnection between decoupled predicted values at various levels.

\section{Experiment}

\subsection{Experimental Setting}

\paragraph{Datasets}  
Experiments are conducted on two commonly-used datasets for oriented object detection, namely DOTA-v1.0~\cite{dota} and HRSC2016~\cite{hrsc2016}.
\textbf{DOTA-v1.0} is a large-scale object detection dataset for optical remote sensing images, which comprises 2,806 images with diverse dimensions, spanning from 800 to 4,000 pixels in width and height. 
The dataset consists of a total of 188,282 individual instances distributed across 15 different classes, and it is partitioned into training, validation, and test sets containing 1,411, 458, and 937 images, respectively.
\textbf{HRSC2016} is an optical remote sensing image dataset designed for ship detection. It comprises 1,680 images with diverse widths and heights, ranging from 339 pixels to 1333 pixels.
The commonly used training, validation, and test sets consist of 436, 181, and 444 images, respectively.

\paragraph{Implementation Details} 

The experimental results are obtained on the MMRotate platform~\cite{mmrotate}. 
We employ the checkpoints of ViT-small/-base~\cite{vit} and HiViT-base~\cite{hivit}, which are all pre-trained using the MAE~\cite{mae} self-supervised strategy. 
We pre-train the ViT-small model and directly utilize the open-sourced checkpoints for the other models, wherein all the Transformer blocks of both the encoder and decoder are fully inherited. 

For a fair comparison, we adopt a similar experimental configuration as used in the benchmark methods~\cite{rvsa, kfiou, orientedrcnn}. 
The performance evaluation on DOTA-v1.0 follows a \textit{multi-scale} setting, where the model is trained on the trainval-set and tested on the test-set. 
In contrast, for the ablation study on DOTA-v1.0, a \textit{single-scale} setting is adopted, where the model is trained on the train-set and tested on the val-set.
All images would be cropped into patches of size 1024$\times$1024 with an overlap of 500/200 pixels in \textit{multi-scale}/\textit{single-scale} setting. 
In the \textit{multi-scale} setting, images are resized by 0.5$\times$, 1.0$\times$, and 1.5$\times$ before undergoing the cropping process, and no scale adjustment is adopted in the \textit{single-scale} setting. 
In the HRSC2016 dataset, images are resized in such a way that the larger dimension of width and height becomes 800, while maintaining their original aspect ratios.

During training, horizontal/vertical flipping and random rotation operations are conducted to increase the scale and diversity of training data.
The model is trained for 12 epochs on DOTA-v1.0 and 36 epochs on HRSC2016.
We adopt the AdamW optimizer~\cite{adam} with an initial learning rate of $1e^{-4}$/$2.5e^{-4}$ for DOTA-v1.0/HRSC2016, a weight decay of $0.05$, and a layer decay of $0.75$/$0.90$ for ViT/HiViT.
All experiments are conducted on 8$\times$A100 GPUs with a batch size of $8$.

\begin{figure}[t]
    \centering
    \includegraphics[width=1\linewidth]{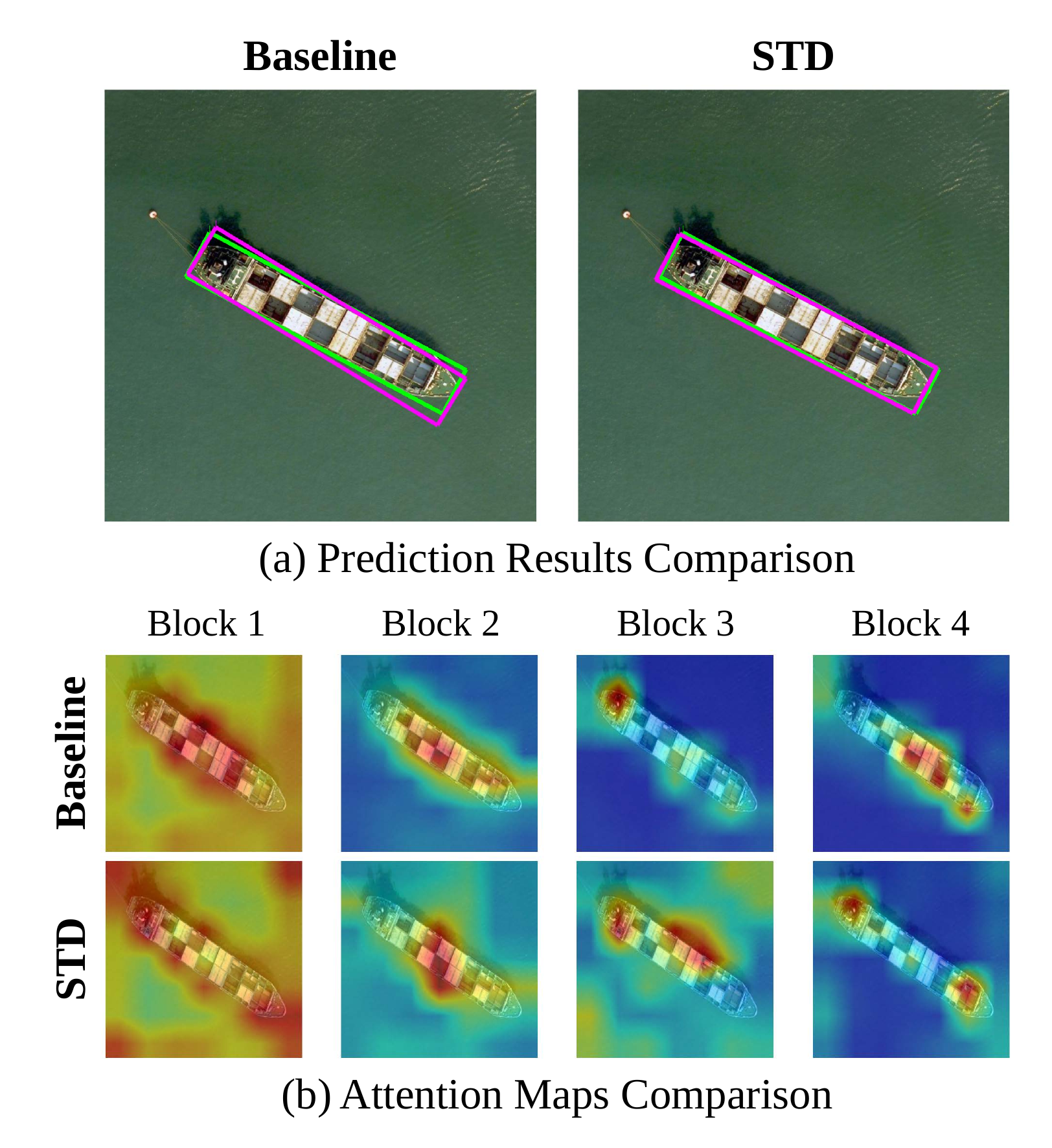}
    \caption{Visualization of attention maps. Compare to the baseline Transformer, the attention maps in STD (bk1 to bk4) exhibit a stronger alignment with the semantic interpretation of the parameter estimated at the respective stage.
    }
    \label{fig:attention_maps}
\end{figure}

\begin{figure*}[t]
    \centering
    \includegraphics[width=1\linewidth]{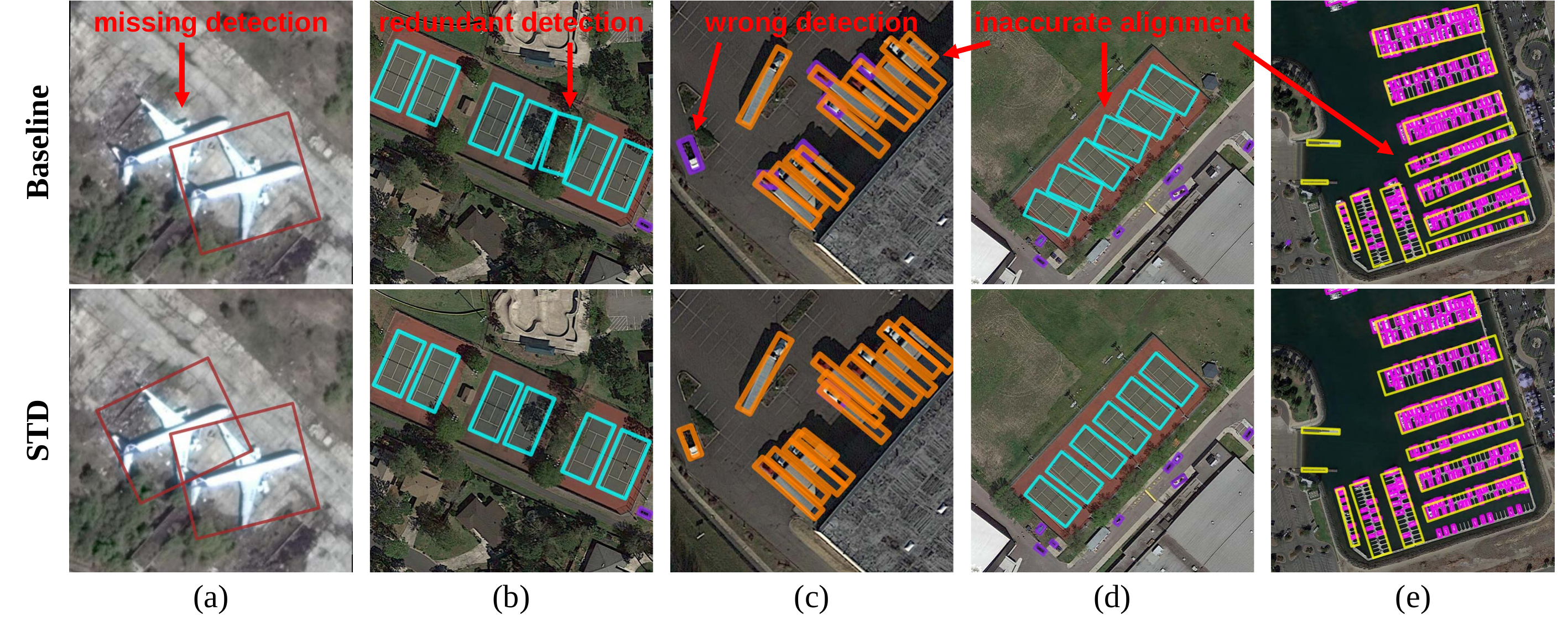}
    \caption{Comparison of detection results. STD demonstrates superior performance in reducing false detections ((a), (b), and (c)), better discerning clustered objects ((c) and (e)), and improving the alignment with oriented objects ((c), (d), and (e)).}
    \label{fig:result_comparison}
\end{figure*}

\subsection{Ablation Study}

\paragraph{Feasibility of Decoupled Parameter Prediction}

Prior to assessing the feasibility of the decoupling approach, we first investigate the performance of bounding box prediction relying solely on a single feature map in various levels.
As shown in Table~\ref{tab:abla_depth}, a consistent enhancement in performance is evident when employing feature maps from deeper layers for bounding box prediction.
This observation suggests that while deep feature maps contribute to improved feature representations for object detection, shallower layers still contain valuable information for bounding box prediction, as their performance is only slightly lower.

We also compare the performance with the decoupled parameter estimation approach. 
As shown in Fig.~\ref{fig:framework}, we used the feature maps from the first block to predict $x$ and $y$, the second block for $\alpha$, the third block for $w$ and $h$, and the final stage for the class score $cls$.
Without the aid of CAMs, the performance of this decoupled configuration is slightly lower than predicting bounding boxes with the feature maps from the third/fourth block by a margin of 0.09\%/0.28\% mAP.
These results provide evidence that suggests the feasibility and potential of designing a decoupled structure.
As previously mentioned, we introduce CAMs to enhance the decoupling process, further reducing the performance gap between decoupled and non-decoupled approaches.

\paragraph{Rationality of Model Design}

To showcase the rationale behind the detailed architecture of STD, we investigate the impact of both the order of parameter decoupling and the location of activation mask integration.
As shown in Table~\ref{tab:abla_order}, the order of decoupling has a significant influence on the performance of STD, and the optimal result is achieved under the configuration $\{x, y\}\rightarrow\alpha\rightarrow\{w, h\}$. 
This phenomenon can be explained by the fact that alternative prediction orders fail to ensure that the RoI could consistently cover the entire foreground object.

\begin{table}[!t]
\centering
\setlength{\tabcolsep}{0.10cm}
\hfill
\begin{subtable}[t]{0.279\textwidth}
\centering
\resizebox{1.0\linewidth}{!}{%
\begin{tabular}[t]{@{}cccc|cc@{}}
\toprule
\multicolumn{4}{c|}{Feature Levels} & \multicolumn{2}{c}{mAP} \\
\midrule
bk1    & bk2      & bk3    & bk4                & -              & +CAMs \\
$\Box$ & -        & -      & $\diamondsuit$     & 69.65          & 71.85 \\
-      & $\Box$   & -      & $\diamondsuit$     & 70.37          & 72.39 \\
-      & -        & $\Box$ & $\diamondsuit$     & 70.88          & 72.27 \\
-      & -        & -      & $\Box\diamondsuit$ & 71.07$\dagger$ & 72.21 \\
$(xy)$ & $\alpha$ & $(wh)$ & $\diamondsuit$     & 70.79          & \textbf{72.78}$\ddagger$ \\
\bottomrule
\end{tabular}
}
\caption{Feature Levels}
\label{tab:abla_depth}
\end{subtable}
\begin{subtable}[t]{0.186\textwidth}
\centering
\resizebox{1.0\linewidth}{!}{%
\begin{tabular}[t]{@{}ccc|c@{}}
\toprule
\multicolumn{3}{c|}{Decoupling Order} &mAP \\
\midrule
$(xy)$  &$(wh)$  &$\alpha$    &72.30  \\
$(wh)$  &$(xy)$  &$\alpha$    &72.26  \\
$(xy)$  &$\alpha$  &$(wh)$    &\textbf{72.78}  \\
$(wh)$  &$\alpha$  &$(xy)$    &72.13  \\
$\alpha$  &$(xy)$  &$(wh)$    &72.03  \\
$\alpha$  &$(wh)$  &$(xy)$    &71.93  \\
\bottomrule
\end{tabular}
}
\caption{Decoupling Order}
\label{tab:abla_order}
\end{subtable}
\caption{Results of diagnostic studies. (a) Comparison of detection accuracy achieved using feature maps from various levels of Transformer block (bk1 to bk4). $\Box$ denotes coupled bounding box prediction and $\diamondsuit$ refers to class score estimation. $\dagger$ indicates the performance of original MAEBBoxHead while $\ddagger$ indicates our STD's. (b) The influence of decoupling order on the overall performance of STD.}
\label{tab:diagnosis}
\end{table}

\begin{table}[!t]
\centering
\begin{tabular}[t]{@{}ccc|c@{}}
\toprule
Detector & Model & BBoxHead & mAP \\
\midrule
Faster RCNN   & ViT-S & MAEBBoxHead & 71.07 \\
Faster RCNN   & ViT-S & STD         & 72.78 \\
\midrule
Oriented RCNN & ViT-S & MAEBBoxHead & 72.41 \\
Oriented RCNN & ViT-S & STD         & 73.43 \\
\bottomrule
\end{tabular}
\caption{Comparison of object detection accuracy achieved by different RoI extraction networks.}
\label{tab:abla_orcnn}
\end{table}

\begin{table*}[t]
\centering
\setlength{\tabcolsep}{0.08cm}
\scriptsize
\resizebox{1.0\linewidth}{!}{%
\begin{tabular}{@{}lcc|ccccccccccccccc|c@{}}
\toprule
Method  &Model  &Pre. 
&PL &BD &BR &GTF &SV &LV &SH &TC &BC &ST &SBF &RA &HA &SP &HC &mAP \\
\midrule
SCRDet~\cite{scrdet} &R101 &IN 
&89.98 &80.65 &52.09 &68.36 &68.36 &60.32 &72.41 &90.85 &87.94 &86.86 &65.02 &66.68 &66.25 &68.24 &65.21 &72.61 \\
Rol Trans.~\cite{roiTransformer} &R50 &IN 
&88.65 &82.60 &52.53 &70.87 &77.93 &76.67 &86.87 &90.71 &83.83 &82.51 &53.95 &67.61 &74.67 &68.75 &61.03 &74.61 \\
Gilding Vertex~\cite{glidingvertex} &R101 &IN 
&89.64 &85.00 &52.26 &77.34 &73.01 &73.14 &86.82 &90.74 &79.02 &86.81 &59.55 &70.91 &72.94 &70.86 &57.32 &75.02 \\
CenterMap~\cite{centermap} &R101 &IN 
&89.83 &84.41 &54.60 &70.25 &77.66 &78.32 &87.19 &90.66 &84.89 &85.27 &56.46 &69.23 &74.13 &71.56 &66.06 &76.03 \\
CSL~\cite{csl} &R152 &IN 
&\textbf{90.25} &85.53 &54.64 &75.31 &70.44 &73.51 &77.62 &90.84 &86.15 &86.69 &69.60 &68.04 &73.83 &71.10 &68.93 &76.17 \\
R$^{3}$Det~\cite{r3det} &R152 &IN 
&89.80 &83.77 &48.11 &66.77 &78.76 &83.27 &87.84 &90.82 &85.38 &85.51 &65.67 &62.68 &67.53 &78.56 &72.62 &76.47 \\
CFA~\cite{cfa} &R152 &IN 
&89.08 &83.20 &54.37 &66.87 &81.23 &80.96 &87.17 &90.21 &84.32 &86.09 &52.34 &69.94 &75.52 &80.76 &67.96 &76.67 \\
SASM~\cite{sasm} &RX101 &IN 
&89.54 &85.94 &57.73 &78.41 &79.78 &84.19 &\textbf{89.25} &90.87 &58.80 &87.27 &63.82 &67.81 &78.67 &79.35 &69.37 &79.17 \\
S$^{2}$ANet~\cite{s2anet} &R50 &IN 
&88.89 &83.60 &57.74 &81.95 &79.94 &83.19 &89.11 &90.78 &84.87 &87.81 &70.30 &68.25 &78.30 &77.01 &69.58 &79.42 \\
ReDet~\cite{redet} &ReR50 &IN 
&88.81 &82.48 &60.83 &80.82 &78.34 &86.06 &88.31 &90.87 &88.77 &87.03 &68.65 &66.90 &79.26 &79.71 &74.67 &80.10 \\
GWD~\cite{gwd} &R152 &IN 
&89.66 &84.99 &59.26 &82.19 &78.97 &84.83 &87.70 &90.21 &86.54 &86.85 &\textbf{73.47} &67.77 &76.92 &79.22 &74.92 &80.23 \\
DEA~\cite{dea} &ReR50 &IN 
&89.92 &83.84 &59.65 &79.88 &80.11 &\textbf{87.96} &88.17 &90.31 &\textbf{88.93} &\textbf{88.46} &68.93 &65.94 &78.04 &79.69 &75.78 &80.37 \\
DODet~\cite{dodet} &R50 &IN 
&89.96 &85.52 &58.01 &81.22 &78.71 &85.46 &88.59 &90.89 &87.12 &87.80 &70.50 &71.54 &82.06 &77.43 &74.47 &80.62 \\
KLD~\cite{kld} &R152 &IN 
&89.92 &85.13 &59.19 &81.33 &78.82 &84.38 &87.50 &89.80 &87.33 &87.00 &72.57 &71.35 &77.12 &79.34 &78.68 &80.63 \\
AOPG~\cite{aopg} &R50 &IN 
&89.88 &85.57 &60.90 &81.51 &78.70 &85.29 &88.85 &90.89 &87.60 &87.65 &71.66 &68.69 &82.31 &77.32 &73.10 &80.66 \\
O-RCNN~\cite{orientedrcnn} &R50 &IN 
&89.84 &85.43 &61.09 &79.82 &79.71 &85.35 &88.82 &90.88 &86.68 &87.73 &72.21 &70.80 &82.42 &78.18 &74.11 &80.87 \\
KFIoU~\cite{kfiou} &Swin-T &IN 
&89.44 &84.41 &\textbf{62.22} &82.51 &80.10 &86.07 &88.68 &\textbf{90.90} &87.32 &88.38 &72.80 &71.95 &78.96 &74.95 &75.27 &80.93 \\
RVSA-O~\cite{rvsa} &ViTAE-B &$\mathcal{M}^\dagger$ 
&88.97 &85.76 &61.46 &81.27 &79.98 &85.31 &88.30 &90.84 &85.06 &87.50 &66.77 &\textbf{73.11} &\textbf{84.75} &\textbf{81.88} &77.58 &81.24 \\
RTMDet-R~\cite{rtmdet} &RTM-L &CO 
&88.01 &\textbf{86.17} &58.54 &82.44 &81.30 &84.82 &88.71 &90.89 &88.77 &87.37 &71.96 &71.18 &81.23 &81.40 &77.13 &81.33 \\
LSKNet~\cite{lsknet} &LSKNet-S &IN 
&89.69 &85.70 &61.47 &\textbf{83.23} &\textbf{81.37} &86.05 &88.64 &90.88 &88.49 &87.40 &71.67 &71.35 &79.19 &81.77 &\textbf{80.86} &81.85 \\
\midrule
\multirow{3}*{KFIoU~\cite{kfiou}} &Swin-B &IN 
&89.26 &86.48 &62.09 &82.86 &79.97 &85.64 &88.47 &90.70 &86.69 &87.54 &71.84 &68.74 &79.62 &81.11 &76.64 &81.18 \\
 &ViT-B   &$\mathcal{M}$ 
&89.32 &84.52 &62.53 &81.86 &81.55 &86.53 &89.00 &90.76 &87.67 &88.29 &67.31 &71.75 &79.63 &80.25 &72.25 &80.88 \\
 &HiViT-B &$\mathcal{M}$ 
&88.71 &86.02 &62.45 &80.24 &80.84 &85.19 &88.47 &90.70 &86.64 &86.37 &67.41 &74.63 &79.04 &80.80 &82.28 &81.32 \\
RVSA-O~\cite{rvsa}   &ViT-B   &$\mathcal{M}^\dagger$ 
&87.63 &85.23 &61.73 &81.11 &80.68 &85.37 &88.26 &90.80 &86.38 &87.21 &67.93 &69.81 &84.06 &81.25 &77.76 &81.01 \\
imTED-O~\cite{imted} &ViT-B   &$\mathcal{M}$ 
&89.41 &84.02 &62.38 &81.05 &81.07 &86.12 &88.47 &90.72 &87.03 &87.18 &68.01 &71.84 &78.87 &79.92 &80.39 &81.10 \\
imTED-O~\cite{imted} &HiViT-B &$\mathcal{M}$ 
&88.24 &84.94 &63.13 &81.70 &80.44 &83.30 &88.18 &90.78 &88.14 &88.23 &71.37 &71.83 &78.77 &80.84 &84.25 &81.61 \\
\midrule
STD-O (ours)         &ViT-B   &$\mathcal{M}$ 
&88.56 &84.53 &62.08 &81.80 &81.06 &85.06 &88.43 &90.59 &86.84 &86.95 &72.13 &71.54 &84.30 &82.05 &78.94 &81.66 \\
STD-O (ours)         &HiViT-B &$\mathcal{M}$ 
&89.15 &85.03 &60.79 &82.06 &80.90 &85.76 &88.45 &90.83 &87.71 &87.29 &\underline{\textbf{73.99}} &71.25 &\underline{\textbf{85.18}} &\underline{\textbf{82.17}} &\underline{\textbf{82.95}} &\underline{\textbf{82.24}} \\
\bottomrule
\end{tabular}
}
\caption{Performance comparison on DOTA-v1.0. Classes: PL-plane; BD-baseball diamond; BR-bridge; GTF-ground track field; SV-small vehicle; LV-large vehicle; SH-ship; TC-tennis court; BC-baseball court; ST-storage tank; SBF-soccer ball field; RA-roundabout; HA-harbor; SP-swimming pool; HC-helicopter. 
Pretraining: IN-supervised pretraining on the ImageNet; CO-supervised pretraining on the MS COCO; $\mathcal{M}$-MAE self-supervised pretraining on the ImageNet; $\mathcal{M}^\dagger$-MAE self-supervised pretraining on the MillionAID~\cite{millionaid}, a large remote sensing dataset including about 1 million images.}
\label{tab:results_dota}
\end{table*}

\paragraph{Adaptability to Different Backbones}

As discussed in the preceding section, as long as the RoI fully covers the foreground object, the activation masks can effectively activate the entire foreground region. 
Hence, our approach is expected to be adaptable to other RoI extraction methodologies. 
As indicated in Table~\ref{tab:abla_orcnn}, the decoupling module of STD also demonstrates strong performance when incorporated into the Oriented RCNN object detector~\cite{orientedrcnn}, which showcases the remarkable generalizability of our method.

\paragraph{Visualization of Attention Maps}
In Figure~\ref{fig:attention_maps}, we present visualizations of the attention maps from different decoder layers of STD and a baseline Transformer model (Rotated Faster RCNN+ViT-S).
In comparison to the baseline Transformer, the attention maps generated by the STD model at each stage exhibit a closer alignment with the semantic meaning of the corresponding predicted parameter.
Specifically, when obtaining the positional information ${x, y}$, the attention tends to concentrate around the center of the object. Following this, the attention becomes more widespread, targeting one end and one edge of the object to capture information about its orientation $\alpha$. Finally, the attention predominantly focuses on both ends of the object, aiming to capture details related to its scale. 
This phenomenon is likely a result of the decoupled bounding box prediction mechanism and the step-wise guidance provided by the activation masks, further confirming the effectiveness of the proposed architectural approach.

\paragraph{Qualitative Comparison}
We also present a qualitative comparison between the results of STD and the baseline Transformer in Figure~\ref{fig:result_comparison}.
STD is capable of mitigating the occurrence of false negatives/positives (as depicted in Figure~\ref{fig:result_comparison}(a), (b)), while also achieving notably improved alignment with oriented foreground objects across different scales (as shown in Figure~\ref{fig:result_comparison}(c), (d), (e)).
This observation highlights the capability of STD in effectively modeling the object orientations without compromising the precision in capturing spatial location and shape information.

\subsection{Performance Comparison}

In this section, we present comprehensive experimental results obtained on the DOTA-v1.0 and HRSC2016 datasets. 
For additional results on the HRSID dataset~\cite{hrsid} and the MS COCO dataset~\cite{mscoco}, please refer to ~\nameref{Appendix3} and ~\nameref{Appendix4}.

\paragraph{DOTA-v1.0}
Table~\ref{tab:results_dota} provides a comprehensive comparison of our method with state-of-the-art approaches on DOTA-v1.0.
All studies have reported their performance in the respective original paper.
We evaluate STD within Oriented RCNN frameworks (STD-O) and both ViT and HiViT models are used for evaluations.
Remarkably, STD achieves new state-of-the-art performance in both frameworks.
When coupled with ViT-B and HiViT-B backbones, STD achieves 81.66\% and 82.24\% mAP, respectively, surpassing the previous best results.

\begin{table}[!t]
\centering
\setlength{\tabcolsep}{0.05cm}
\footnotesize 
\begin{tabular}{@{}l|cccc@{}}
\toprule
Method & Model & Pre. & mAP(07) & mAP(12) \\
\midrule
CenterMap~\cite{centermap} &R101 &IN &- &92.70 \\
RoI Trans.~\cite{roiTransformer} &R101 &IN &86.20 &- \\
Gilding Vertex~\cite{glidingvertex} &R101 &IN &88.20 &- \\
R$^{3}$Det~\cite{r3det} &R101 &IN &89.26 &96.01 \\
DAL~\cite{dal} &R101 &IN &89.77 &- \\
GWD~\cite{gwd} &R101 &IN &89.85 &97.37 \\
S$^2$ANet~\cite{s2anet} &R101 &IN &90.17 &95.01 \\
AOPG~\cite{aopg} &R101 &IN &90.34 &96.22 \\
ReDet~\cite{redet} &ReR50 &IN &90.46 &97.63 \\
O-RCNN~\cite{orientedrcnn} &R101 &IN &90.50 &97.60 \\
RTMDet-R~\cite{rtmdet} &RTM-L &CO &90.60 &97.10 \\
\midrule
STD-O (ours)  &ViT-B   &$\mathcal{M}$ &\underline{\textbf{90.67}} &\underline{\textbf{98.55}} \\
STD-O (ours)  &HiViT-B &$\mathcal{M}$ &90.63 &98.20 \\
\bottomrule
\end{tabular}
\caption{Comparison of performance on HRSC2016.}
\label{tab:results_hrsc}
\end{table}

\paragraph{HRSC2016}
In Table~\ref{tab:results_hrsc}, the experimental results demonstrate that the STD method outperforms all other methods, achieving an impressive mAP of 90.67\% and 98.55\% under the PASCAL VOC 2007~\cite{pascalvoc} and VOC 2012 metrics, respectively.

\paragraph{Computational Cost.} STD has a small computational cost overhead. While the baseline detector with HiViT-B takes around 288ms on average to process an image using A100 GPUs, the overall STD detector requires an average processing time of 315ms.

\section{Conclusion}
This paper introduces Spatial Transform Decoupling (STD), an oriented object detection method that separates the parameter prediction process into multiple disentangled stages.
Such a decoupled process is further enhanced by incorporating cascaded activation masks, which introduce dense guidance into the self-attention mechanism. 
Extensive experiments have demonstrated the effectiveness of STD on multiple popular benchmarks.
To the best of our knowledge, STD is a pioneering method that tackles oriented object detection in remote sensing with a structural perspective. 
Notably, the Transformer-based nature of STD enables seamless integration with various advanced pre-trained models, providing significant benefits to the research community.

%

\bibliography{aaai24}

\clearpage
\newpage

\section{A. Appendix}

\subsection{A.1. Detailed Derivation of the Spatial Transform}
\label{Appendix1}

\begin{figure}[hb]
    \centering
    \includegraphics[width=1.0\linewidth]{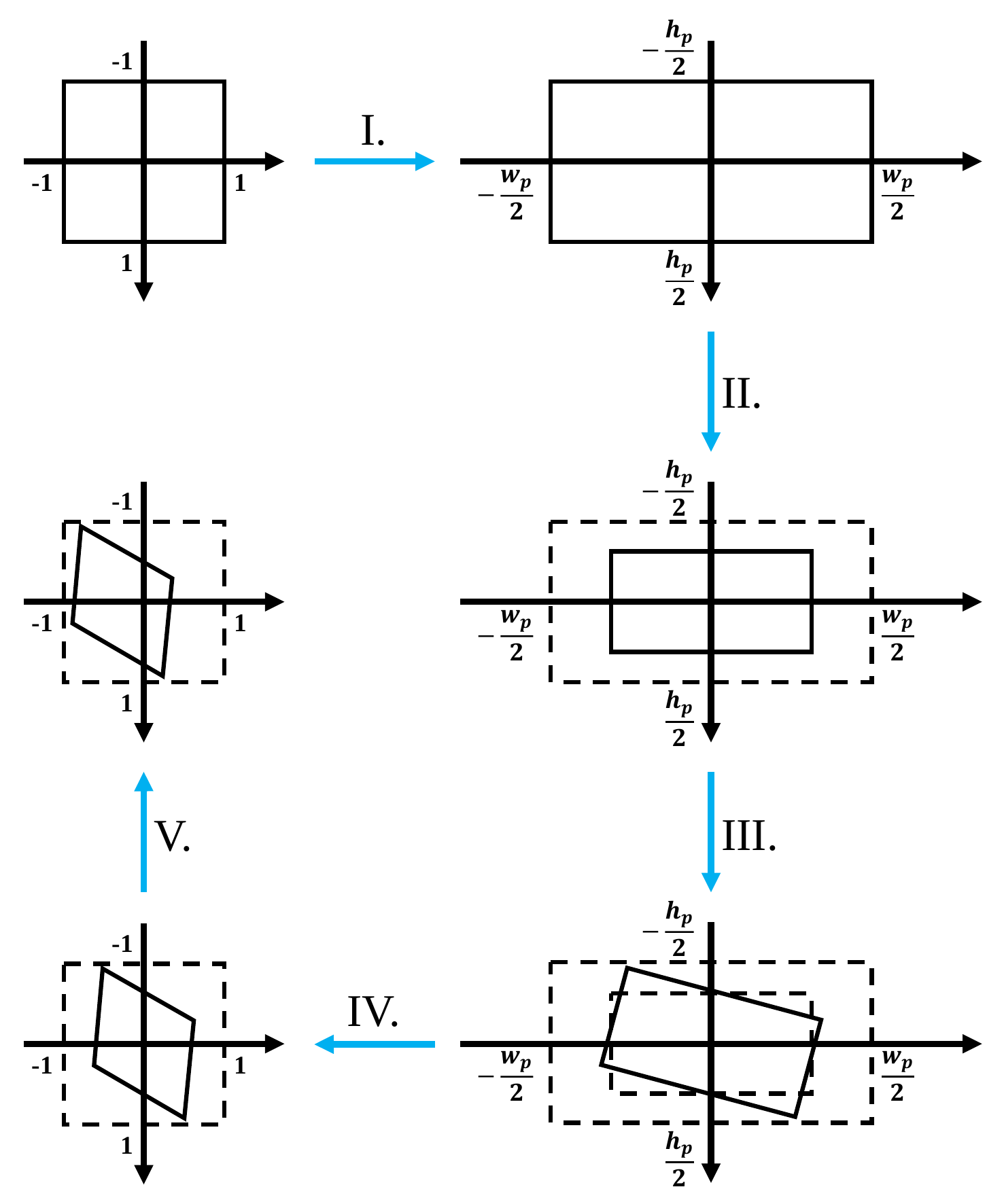}
    \caption{Illustration of the decomposed five steps of the affine transformation.}
    \label{fig:appendix_figa1}
\end{figure}

In the main submission, we have directly presented the affine transformation as:
\begin{equation}
\label{eq:appendix_affine_transform}
    \resizebox{1.0\hsize}{!}{$%
    \left(\begin{array}{c}
                x^{'}  \\
                y^{'} 
            \end{array}\right)\\
    =\left(\begin{array}{cc}
                \cos d\alpha \cdot e^{dw} & -\sin d\alpha \cdot e^{dh} \cdot \frac{h_p}{w_p} \\
                \sin d\alpha \cdot e^{dw} \cdot \frac{w_p}{h_p} & \cos d\alpha \cdot e^{dh}
            \end{array}\right)
      \left(\begin{array}{c}
                x  \\
                y 
            \end{array}\right) \\
    +\left(\begin{array}{c}
                2\cdot dx \\
                2\cdot dy
            \end{array}\right). \\
    $}%
\end{equation}
As illustrated in Figure~\ref{fig:actv_mask}, the activation mask could be produced by applying the affine transformation in Eq.(\ref{eq:appendix_affine_transform}) to a matrix $\boldsymbol{AM}$ with all elements set to 1.

In this section, we will provide a detailed and comprehensive derivation of this transformation.
In agreement with the main submission, let us denote the central coordinates and dimensions of a proposal as $(x_p, y_p, w_p, h_p)$, and the predicted values obtained by the model as $(dx, dy, dw, dh, d\alpha)$.
We place the feature map within a rectangular coordinate system $(x, y)$ and its four vertices located at $(-1, -1)$, $(1, -1)$, $(1, 1)$, and $(-1, 1)$~\footnote{This positioning scheme is in fact tailored to align with the settings of PyTorch's `affine\_grid/grid\_sample' functions.}.
Practically, the affine transformation from the square feature map to the bounding box (within the square feature map) is decomposed into five distinct steps (see Figure~\ref{fig:appendix_figa1}):
\begin{enumerate}[I.]
    \item Reshape the square feature map to the original rectangular proposal;
    \item Scale the height and width of the rectangular proposal;
    \item Rotate the transformed proposal;
    \item Perform the inverse transformation of the Step I.;
    \item Perform the spatial translation.
\end{enumerate}
These five transformations could be represented by five matrices: $M_{f{\rightarrow}p}$, $M_{wh}$, $M_{a}$, $M_{p{\rightarrow}f}$, and $M_{xy}$, and then the overall transformation could be written as:
\begin{equation}
\begin{aligned}
&\left(\begin{array}{c}
     x^{'}  \\
     y^{'} 
\end{array}\right)\\
&=M_{p{\rightarrow}f}\cdot M_{\alpha} \cdot M_{wh} \cdot M_{f{\rightarrow}p} \cdot \left(\begin{array}{c}
     x  \\
     y 
\end{array}\right) + M_{xy}, \\
&=\left(\begin{array}{cc}
    \frac{2}{w_p} & 0 \\
    0 & \frac{2}{h_p}
\end{array}\right) \left(\begin{array}{cc}
    cos\alpha & -sin\alpha \\
    sin\alpha & cos\alpha
\end{array}\right) \left(\begin{array}{cc}
    e^{dw} & 0 \\
    0 & e^{dh}
\end{array}\right) \\
& \left(\begin{array}{cc}
    \frac{w_p}{2} & 0 \\
    0 & \frac{h_p}{2} 
\end{array}\right)  
 \left(\begin{array}{c}
    x  \\
    y 
\end{array}\right) + \left(\begin{array}{c}
    2\cdot dx \\
    2\cdot dy
\end{array}\right), \\
&=\left(\begin{array}{cc}
    cos\alpha \cdot e^{dw}  & -sin\alpha \cdot e^{dh} \cdot \frac{h_p}{w_p} \\
    sin\alpha \cdot e^{dw} \cdot \frac{w_p}{h_p}& cos\alpha \cdot e^{dh}
\end{array}\right)  
 \left(\begin{array}{c}
    x  \\
    y 
\end{array}\right) \\
&+ \left(\begin{array}{c}
    2\cdot dx \\
    2\cdot dy
\end{array}\right)  \\
\end{aligned}
\label{eq:affine_conversion_aligned}
\end{equation}
which is exactly the same expression as in Eq.(\ref{eq:appendix_affine_transform}).

\subsection{A.2. Ablation Study on Detailed Network Design}
\label{Appendix2}

Ablation studies are conducted to further validate two detailed aspects of the network design.

\paragraph{Position for Activation Mask Integration}
As mentioned in the main submission, activation masks ($\boldsymbol{AM}$) are computed and integrated into the self-attention blocks of the Transformer decoder to provide dense guidance. 
However, there are multiple options for how to incorporate $\boldsymbol{AM}$, and the performance of each setting is reported in Table~\ref{tab:appendix_tab1}.
It is evident that while applying $\boldsymbol{AM}$ to both $\boldsymbol{Q}$ and $\boldsymbol{K}$ can achieve a similar (though slightly inferior) performance, the best result is obtained by exclusively multiplying $\boldsymbol{AM}$ with $\boldsymbol{V}$, which also indeed offers a lower computational cost.

\paragraph{Number of Auxiliary Convolutional Layers}
Experiments are also conducted to investigate the impact of the number of convolutional layers integrated into the decoupling module.
Intuitively, incorporating convolutional layers before the parameter estimation network can elevate the inherent expressiveness of extracted feature maps, which contributes to an enhanced accuracy in parameter regression.
To comprehensively investigate the potential impacts of introducing convolutional layers, we conducted a thorough series of tests, ranging from branch designs without convolutional layers to the architecture proposed in the main submission. Throughout these tests, both decoupled parameter prediction and cascaded activation masks (CAMs) were employed.

\begin{table}[!t]
\centering
\begin{tabular}[t]{ccc|c}
\toprule
\multicolumn{3}{c|}{$\boldsymbol{AM}$ $\bigodot$ \textit{with}} & \multirow{2}*{mAP} \\
$\boldsymbol{Q}$ & $\boldsymbol{K}$ & $\boldsymbol{V}$ \\
\midrule
\checkmark &            &            & 71.90 \\
\checkmark & \checkmark &            & 72.71 \\
\checkmark & \checkmark & \checkmark & 72.33 \\
           & \checkmark & \checkmark & 72.51 \\
           &            & \checkmark & \textbf{72.78} \\
\bottomrule
\end{tabular}
\caption{Performance comparison of different positions for activation mask integration.}
\label{tab:appendix_tab1}
\end{table}

\begin{table}[!t]
\centering
\begin{tabular}[t]{ccc|c}
\toprule
$(xy)$ & $\alpha$ & $(wh)$ & mAP   \\
\midrule
0      & 0        & 0      & 71.84 \\
\midrule
1      & 0        & 0      & 72.00 \\
0      & 1        & 0      & 71.90 \\
0      & 0        & 1      & 71.56 \\
\midrule
2      & 0        & 0      & 72.28 \\
0      & 2        & 0      & 72.05 \\
0      & 0        & 2      & 71.89 \\
\midrule
3      & 0        & 0      & 72.44 \\
0      & 3        & 0      & 71.94 \\
0      & 0        & 3      & 71.57 \\
\midrule
3      & 2        & 1      & \textbf{72.78} \\
2      & 2        & 2      & 71.96 \\
1      & 2        & 3      & 71.19 \\
\bottomrule
\end{tabular}
\caption{Performance comparison of the different configurations of auxiliary convolutional layers.}
\label{tab:appendix_tab2}
\end{table}

According to results listed in Table~\ref{tab:appendix_tab2}, we have made the following observations:
(1) Given an equal number of additional convolutional layers, the decision to apply them to feature maps in shallower layers seems more appropriate than incorporating them into deeper ones. This choice is motivated by the inherent limitations in the representational capacity of shallow feature maps, and the introduction of convolutional layers effectively mitigates this constraint.
(2) Excessively augmenting deep feature maps with additional convolutional layers is found to have a more negative impact than positive. This is attributed to the fact that deep feature maps inherently contain refined and compact high-level semantic information. The introduction of additional convolutional layers tends to introduce disturbances rather than providing benefits.
Taking both these factors into consideration, we can deduce that when there is a possibility to include additional convolutional layers, it would be more beneficial to allocate these resources primarily towards feature maps in the shallower stages. This conclusion is further supported by the experimental findings in Table~\ref{tab:appendix_tab2}. The potential existence of more efficient auxiliary structures for feature extraction, such as specialized convolutional kernel designs, remains a topic for future investigation.

\subsection{A.3. Experiment Results on HRSID}
\label{Appendix3}

\begin{table}[!t]
\centering
\setlength{\tabcolsep}{0.15cm}
\footnotesize 
\begin{tabular}{l|cccc}
\toprule
Method & Model &Pre. & mAP (ins.) & mAP (offs.) \\
\midrule
\multicolumn{5}{l}{\textit{Faster RCNN framework}~\cite{fasterrcnn}} \\
Faster RCNN   &Swin-T  &IN  &50.63 &89.24 \\
Faster RCNN   &Swin-S  &IN  &51.28 &89.47 \\
Faster RCNN   &Swin-B  &IN  &52.22 &89.26 \\
imTED-F       &ViT-S   &MAE &46.00 &88.95 \\
imTED-F       &ViT-B   &MAE &49.01 &88.92 \\
imTED-F       &HiViT-B &MAE &50.67 &89.30 \\
STD-F         &ViT-S   &MAE &50.31 &89.80 \\
STD-F         &ViT-B   &MAE &\textbf{52.96} &\textbf{89.85} \\
STD-F         &HiViT-B &MAE &51.54 &89.46 \\
\midrule
\multicolumn{5}{l}{\textit{Oriented RCNN framework}~\cite{orientedrcnn}} \\
Oriented RCNN &Swin-T  &IN  &56.61 &90.69 \\
Oriented RCNN &Swin-S  &IN  &56.50 &90.64 \\
Oriented RCNN &Swin-B  &IN  &57.16 &90.75 \\
imTED-O       &ViT-S   &MAE &53.99 &90.48 \\
imTED-O       &ViT-B   &MAE &54.86 &90.66 \\
imTED-O       &HiViT-B &MAE &55.94 &90.63 \\
STD-O         &ViT-S   &MAE &55.35 &90.75 \\
STD-O         &ViT-B   &MAE &\underline{\textbf{57.59}} &\underline{\textbf{90.84}} \\
STD-O         &HiViT-B &MAE &56.35 &90.80 \\
\bottomrule
\end{tabular}
\caption{Comparison of performance on HRSID dataset.}
\label{tab:appendix_hrsid}
\end{table}

We have also conducted performance evaluations on the HRSID dataset~\cite{hrsid} for oriented object detection. The majority of existing studies lack performance reports on this dataset. Consequently, all experiments in this section were carried out on ourselves using 8$\times$A100 GPUs.

HRSID is a high-resolution synthetic aperture radar (SAR) image dataset specifically curated for ship object detection.
It comprises 5,604 high-resolution SAR images and contains a total of 16,951 ship instances. 
All images in HRSID are scaled to be of size 800$\times$800. 
Experiments are conducted under both Faster R-CNN~\cite{fasterrcnn} and Oriented R-CNN~\cite{orientedrcnn} detectors for a fair comparison, and the widely acknowledged Swin Transformer~\cite{swin} is used as the backbone network in the baseline method. 
Hyperparameters such as data augmentation and initial learning rate are the same as those used on DOTA-v1.0~\cite{dota}, and the model is trained for 36 epochs on HRSID.
Results are reported in Table~\ref{tab:appendix_hrsid}.

We discover that employing the imTED structure~\cite{imted} alone (i.e. MAEBBoxHead only) struggles to surpass the baseline performance. However, our architecture exhibits noticeable performance enhancements over imTED and ultimately outperforms the baseline.
Furthermore, we notice that the ViT-B architecture~\cite{vit} surprisingly outperforms the theoretically superior HiViT-B architecture~\cite{hivit}. This observation contradicts the outcomes on the DOTA-v1.0 dataset but is aligned with the performance on the HRSC2016 dataset~\cite{hrsc2016}. We hypothesize that this deviation might be attributed to the dataset's relatively smaller size, potentially leading to instances of overfitting.

\begin{table*}[t]
\centering
\setlength{\tabcolsep}{0.09cm}
\scriptsize
\begin{tabular}{c|ccccccccc ccccccccc ccccccccc }
\toprule
Method & 1 & 2 & 3 & 4 & 5 & 6 & 7 & 8 & 9 & 10 & 11 & 12 & 13 & 14 & 15 & 16 & 17 & 18 & 19 & 20 & 21 & 22 & 23 & 24 & 25 & 26 & 27 \\
\midrule
imTED   & 55.3 & 32.5 & 45.5 & 44.7 & 69.6 & 68.4 & 66.6 & 38.1 & 29.7 & 26.7 & 69.6 & 64.7 & 46.9 & 29.1 & 37.6 & 71.0 & 67.1 & 58.2 & 52.9 & 61.4 & 68.0 & 75.3 & 68.2 & 68.0 & 14.9 & 41.3 & 14.6 \\
STD-HBB & 55.1 & 33.5 & 45.3 & 45.5 & 68.1 & 68.2 & 66.6 & 37.3 & 30.2 & 26.5 & 70.7 & 64.6 & 46.3 & 28.5 & 38.1 & 70.5 & 66.4 & 58.7 & 53.2 & 60.9 & 67.1 & 74.9 & 69.3 & 69.5 & 15.7 & 41.4 & 15.2 \\
\midrule
\midrule
Method & 28 & 29 & 30 & 31 & 32 & 33 & 34 & 35 & 36 & 37 & 38 & 39 & 40 & 41 & 42 & 43 & 44 & 45 & 46 & 47 & 48 & 49 & 50 & 51 & 52 & 53 & 54 \\
\midrule
imTED   & 35.6 & 42.9 & 68.9 & 27.4 & 39.5 & 44.5 & 42.7 & 34.8 & 38.8 & 53.1 & 44.8 & 49.5 & 40.2 & 37.7 & 42.7 & 39.1 & 19.4 & 21.6 & 43.7 & 26.5 & 18.5 & 38.0 & 29.6 & 25.5 & 22.7 & 37.8 & 54.9 \\
STD-HBB & 34.4 & 43.7 & 67.3 & 27.0 & 41.7 & 44.5 & 43.8 & 33.3 & 37.9 & 55.3 & 43.4 & 49.2 & 39.8 & 36.6 & 43.2 & 40.4 & 21.7 & 22.3 & 43.4 & 25.6 & 19.6 & 36.3 & 29.5 & 24.1 & 22.8 & 37.3 & 56.1 \\
\midrule
\midrule
Method & 55 & 56 & 57 & 58 & 59 & 60 & 61 & 62 & 63 & 64 & 65 & 66 & 67 & 68 & 69 & 70 & 71 & 72 & 73 & 74 & 75 & 76 & 77 & 78 & 79 & 80 & \textbf{AP} \\
\midrule
imTED   & 48.5 & 40.9 & 31.9 & 49.0 & 28.1 & 51.0 & 30.3 & 63.3 & 57.2 & 63.2 & 61.3 & 37.0 & 54.0 & 37.9 & 57.5 & 35.9 & 47.0 & 38.4 & 56.8 & 16.6 & 48.5 & 38.1 & 39.9 & 51.6 & 10.0 & 34.4 & \textbf{44.2} \\
STD-HBB & 49.3 & 40.0 & 31.3 & 48.1 & 29.3 & 50.5 & 29.4 & 62.4 & 58.2 & 62.4 & 59.4 & 37.5 & 53.9 & 37.0 & 60.5 & 37.4 & 43.5 & 38.4 & 58.8 & 16.9 & 49.9 & 39.2 & 39.2 & 50.6 & 11.1 & 31.7 & \textbf{44.2} \\
\bottomrule
\end{tabular}
\caption{Comparison of performance on MS COCO.
\textbf{Classes}.
1: 'person'.
2: 'bicycle'.
3: 'car'.
4: 'motorcycle'.
5: 'airplane'.
6: 'bus'.
7: 'train'.
8: 'truck'.
9: 'boat'.
10: 'traffic light'.
11: 'fire hydrant'.
12: 'stop sign'.
13: 'parking meter'.
14: 'bench'.
15: 'bird'.
16: 'cat'.
17: 'dog'.
18: 'horse'.
19: 'sheep'.
20: 'cow'.
21: 'elephant'.
22: 'bear'.
23: 'zebra'.
24: 'giraffe'.
25: 'backpack'.
26: 'umbrella'.
27: 'handbag'.
28: 'tie'.
29: 'suitcase'.
30: 'frisbee'.
31: 'skis'.
32: 'snowboard'.
33: 'sports ball'.
34: 'kite'.
35: 'baseball bat'.
36: 'baseball glove'.
37: 'skateboard'.
38: 'surfboard'.
39: 'tennis racket'.
40: 'bottle'.
41: 'wine glass'.
42: 'cup'.
43: 'fork'.
44: 'knife'.
45: 'spoon'.
46: 'bowl'.
47: 'banana'.
48: 'apple'.
49: 'sandwich'.
50: 'orange'.
51: 'broccoli'.
52: 'carrot'.
53: 'hot dog'.
54: 'pizza'.
55: 'donut'.
56: 'cake'.
57: 'chair'.
58: 'couch'.
59: 'potted plant'.
60: 'bed'.
61: 'dining table'.
62: 'toilet'.
63: 'tv'.
64: 'laptop'.
65: 'mouse'.
66: 'remote'.
67: 'keyboard'.
68: 'cell phone'.
69: 'microwave'.
70: 'oven'.
71: 'toaster'.
72: 'sink'.
73: 'refrigerator'.
74: 'book'.
75: 'clock'.
76: 'vase'.
77: 'scissors'.
78: 'teddy bear'.
79: 'hair drier'.
80: 'toothbrush'.}
\label{tab:appendix_coco}
\end{table*}

\subsection{A.4. Experiment Results on MS COCO}
\label{Appendix4}

We have also conducted an extrapolation of the STD method on a widely used general object detection dataset, $i.e.$, MS COCO~\cite{mscoco}. 
Specifically, we modify STD by removing the angle prediction branch while retaining the CAMs structure (named `STD-HBB'). Nevertheless, in this scenario, the activation masks adopt the form of a horizontal rectangle instead of a rotated quadrilateral. 
According to the results reported in Table~\ref{tab:appendix_coco}, there is no significant disparity in performance between STD-HBB and imTED~\cite{imted} on ViT-small 1$\times$~\cite{vit}.

We conducted a comprehensive analysis of the Average Precision (AP) for each category and observed that these two methods exhibit distinct strengths in different specific classes.
The most significant improvement was observed in the 'microwave' category (+3.0\% AP), while the most noticeable decrease was observed in the 'toaster' category (-3.5\% AP). Notably, although these two categories exhibit remarkably similar external appearances, they result in vastly different detection accuracies when assessed using different methods.

This may be caused by the intrinsic difference between the decoupling property of the representation of general objects in MS COCO and oriented objects in remote sensing images. 
Specifically, general objects are typically characterized by irregular shapes and are commonly positioned at consistent angles within typical scenes (e.g., people and vehicles standing upright on the ground). Hence, assigning orientation information to such objects might not hold much practical value. Additionally, general objects are usually defined using horizontal bounding boxes ($x$, $y$, $w$, $h$), and these parameters inherently become intertwined when the image undergoes rotation.
In contrast, objects in remote sensing scenarios are typically observed from an aerial perspective, which allows them to be present at various angles, making them inherently oriented. As a result, these objects are often described using oriented bounding boxes ($x$, $y$, $w$, $h$, $\alpha$). In particular, it should be emphasized that while angle ($\alpha$) and the position ($x$, $y$) can vary as the images are rotated, the shape parameters ($w$, $h$) remain invariant to rotation.

Therefore, this underscores the fact that our proposition of STD from the perspective of decoupled angle predictions is indeed more aligned with the characteristics of oriented object detection tasks. 
However, even though applying the STD structure directly to general object detection tasks goes against our initial design intention, it still manages to achieve performance comparable to the baseline model designed for general object detection. This demonstrates the generalization capability of the proposed STD framework.

\subsection{A.5. Qualitative Results on DOTA-v1.0}
\label{Appendix5}

\begin{figure*}[t]
    \centering
    \includegraphics[width=0.98\linewidth]{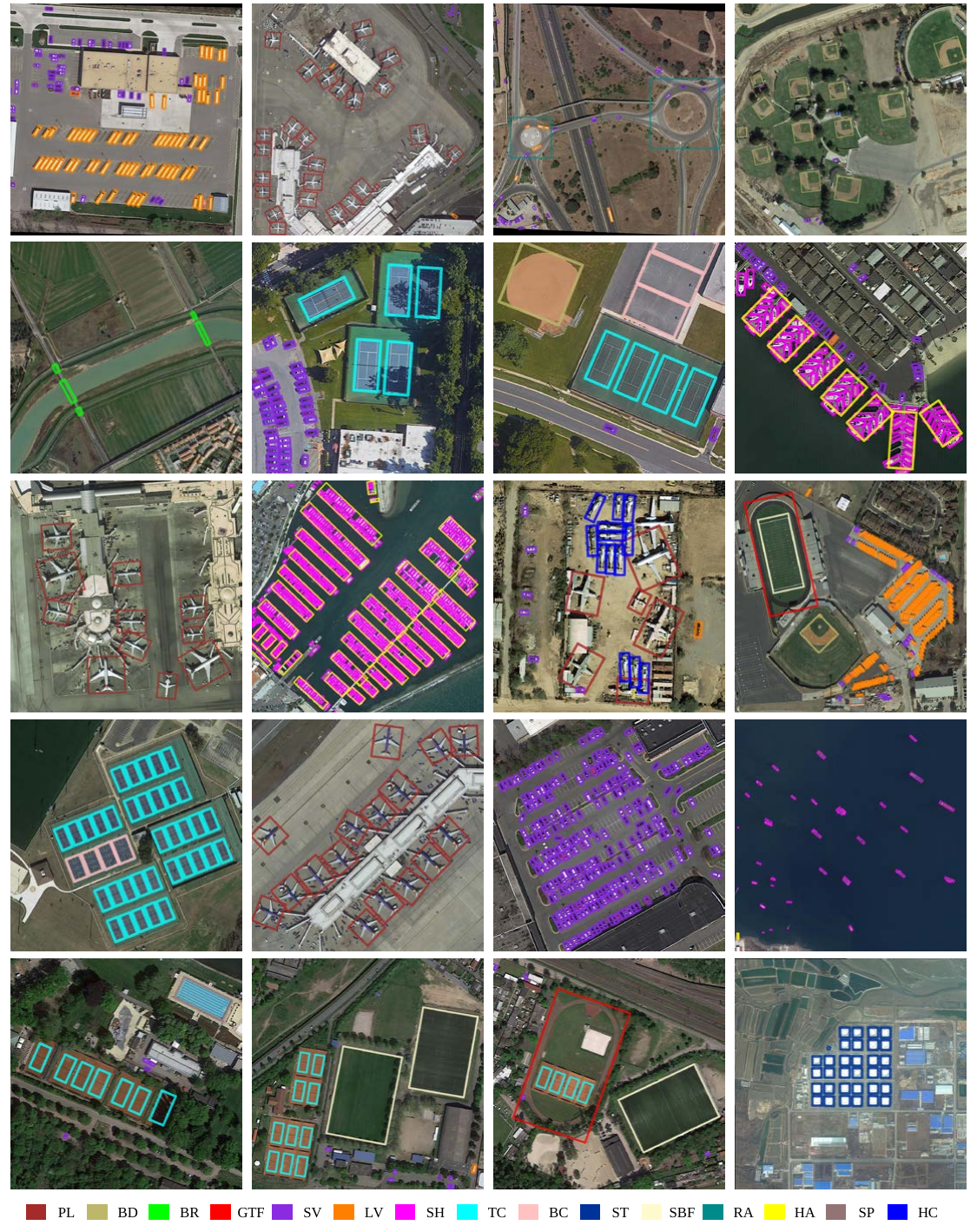}
    \caption{Qualitative detection results on the DOTA-v1.0 dataset. Category labels are the same as in Table~\ref{tab:results_dota}.}
    \label{fig:appendix_figa2}
\end{figure*}

Qualitative results obtained on the DOTA-v1.0 dataset are shown in Figure~\ref{fig:appendix_figa2}.
Although the input images cover a wide range of scenes containing objects in different categories with various scales, the proposed STD could accurately predict oriented bounding boxes well-aligned with the target, demonstrating the effectiveness of STD in detecting oriented objects in remote sensing scenarios.

\end{document}